\documentclass[conference]{IEEEtran}
\IEEEoverridecommandlockouts
\usepackage{cite}
\usepackage{amsmath,amssymb,amsfonts}
\usepackage{algorithmic}
\usepackage{graphicx}
\usepackage{flafter}
\usepackage{textcomp}
\usepackage{xcolor}
\usepackage{booktabs}
\usepackage{multirow}
\def\BibTeX{{\rm B\kern-.05em{\sc i\kern-.025em b}\kern-.08em
    T\kern-.1667em\lower.7ex\hbox{E}\kern-.125emX}}
\begin{document}

\title{Uncertainty-Guided Sparse Refinement for Action Chunking Transformer Policies}

\author{
\IEEEauthorblockN{Chenyang Wang$^{1,\dagger}$, Yuntian Wang$^{1,\dagger}$, Xiaoxiong Yang$^{2}$, Dingde Jiang$^{2}$, Siao Liu$^{1,\ddagger}$, Yang Liu$^{3}$}
\IEEEauthorblockA{$^{1}$Soochow University, China, $^{2}$University of Electronic Science and Technology of China, China, $^{3}$Tongji University, China}
\IEEEauthorblockA{$^{\dagger}$Equal contribution, $^{\ddagger}$Corresponding author}
}

\maketitle
\vspace{-1.6em}

\begin{abstract}
Learning chunk-based visuomotor policies for long-horizon robot manipulation remains challenging. Recent action-chunking methods have shown promising performance by predicting temporally extended action sequences. However, their failures are often dominated by prediction errors at a small number of critical timesteps rather than uniformly poor predictions across the entire action chunk, making uniform refinement inefficient and insufficiently targeted. To address this bottleneck, we propose \textbf{Uncertainty-Guided Refinement (UGR)}, a sparse refinement framework for chunk-based visuomotor policies. Specifically, UGR follows a coarse-to-refine design: it first predicts a full action chunk, estimates per-step temporal uncertainty from the coarse hidden states, and applies residual correction only to the most uncertain timesteps selected by a binary mask. The uncertainty branch is decoupled from the coarse action predictor, enabling clean attribution of the refinement gains to uncertainty-guided correction rather than additional predictor capacity. Extensive experiments on five dual-arm manipulation tasks from the RoboTwin benchmark show that UGR achieves the best success rate on four tasks, improves over the ACT baseline by up to 13\% absolute, and outperforms both full-chunk and position-agnostic block refinement in ablation studies.
\end{abstract}

\begin{figure}[t]
\centering
\includegraphics[width=0.98\columnwidth]{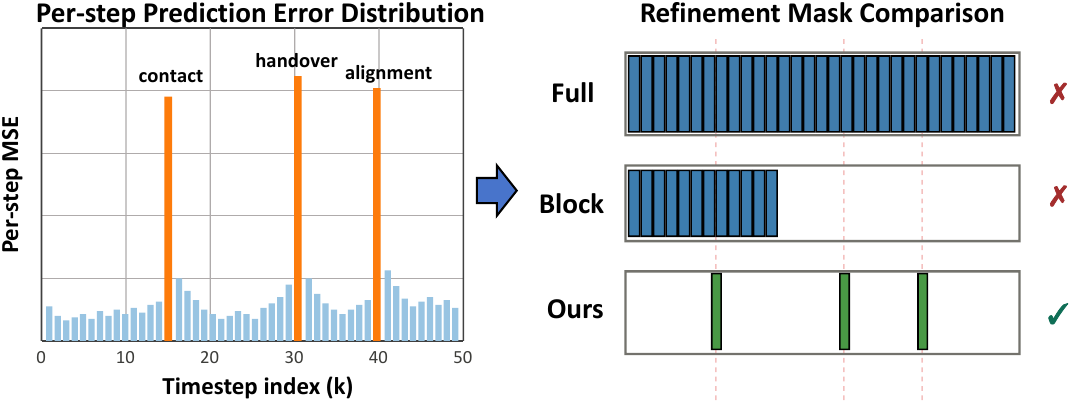}
\vspace{-0.6em}
\caption{Uncertainty-guided sparse refinement targets temporally uncertain steps for residual correction, instead of refining the full action chunk or a position-agnostic block.}
\label{fig:teaser}
\vspace{-1.0em}
\end{figure}

\section{Introduction}
\label{sec:intro}

Recent discussions on embodied intelligence and embodied multimedia highlight growing interest in agents that connect perception with physical interaction~\cite{zuo2026embodied}. Within this setting, visuomotor imitation learning maps visual observations to robot actions from demonstrations and has become a central paradigm for robot manipulation~\cite{mandlekar2022matters,zhao2023act,chi2025diffusion,florence2022ibc}. A fundamental difficulty is \emph{compounding errors}: small step-wise prediction errors accumulate over long horizons and drive the policy into out-of-distribution states~\cite{ross2011reduction}. Chunk-based action prediction and temporal action abstraction mitigate this issue by predicting or executing temporally extended actions~\cite{zhao2023act,chi2025diffusion,belkhale2023hydra}.

The Action Chunking Transformer (ACT)~\cite{zhao2023act} is a representative chunk-based policy for bimanual manipulation. Related methods, including Diffusion Policy~\cite{chi2025diffusion}, Behavior Transformer~\cite{shafiullah2022bet}, VQ-BeT~\cite{lee2024vqbet}, PerAct~\cite{shridhar2023peract}, and OpenVLA~\cite{kim2024openvla}, further demonstrate the effectiveness of temporally extended or high-capacity visuomotor prediction, but can still fail on challenging long-horizon tasks.
More broadly, related robot learning work has studied robustness and skill reuse from complementary perspectives, including conflict-aware augmentation for visual reinforcement learning generalization, diffusion-based skill denoising for robotic manipulation, and primitive-level prompting for lifelong skill reuse~\cite{liu2023cg2a,liu2024diffskill,yao2025primitive}.

We observe that such failures are often not caused by uniformly poor predictions across the entire chunk. Figure~\ref{fig:teaser} illustrates this localized-error motivation and the corresponding sparse refinement mask. Instead, errors often concentrate around a small number of critical timesteps, especially near contact-rich transitions such as alignment, handover, or mechanism-triggering events. Our ablation results support this view: targeting the top 30\% most uncertain timesteps outperforms both full-chunk and random-block refinement (Table~\ref{tab:ablation}). This suggests that improving ACT policies does not necessarily require uniformly stronger global prediction, but rather targeted refinement of the most error-prone local regions.

Prior visuomotor sequence models typically improve action chunks globally, either through stronger generative backbones or additive residual correction, without distinguishing between easy and hard timesteps within a chunk. Residual policy learning~\cite{silver2018residual} illustrates additive correction, while Diffusion Policy~\cite{chi2025diffusion} performs iterative global denoising without per-step selectivity. Less explored is whether refinement capacity can be allocated \emph{within} a chunk according to timestep-level difficulty. This gap motivates a sparse, data-dependent refinement mechanism that identifies and corrects only the most difficult timesteps.

To this end, we propose an uncertainty-guided sparse refinement framework for ACT policies. The key idea is a coarse-to-refine design: the policy first predicts a full action chunk, then identifies unreliable timesteps via learned temporal uncertainty, and finally applies residual correction only at those selected locations. We evaluate the framework on five dual-arm manipulation tasks from the RoboTwin benchmark~\cite{mu2024robotwin} under a unified ACT training and evaluation protocol, comparing it against baselines and ACT-family refinement variants.

The main contributions of this paper are as follows:
\begin{itemize}
    \item We propose an uncertainty-guided sparse refinement framework for ACT policies that uses learned temporal uncertainty to select where residual correction should be applied within an action chunk, concentrating refinement capacity on the most uncertain timesteps.
    \item We introduce a decoupled uncertainty learning design, where uncertainty is estimated from detached coarse states and used purely as a temporal selection signal for refinement.
    \item We evaluate the proposed method on five RoboTwin manipulation tasks under a unified ACT training and evaluation protocol, and preliminary results show improved or competitive performance over representative baselines and ACT-family refinement variants.
\end{itemize}

\begin{figure*}[t]
\centering
\includegraphics[width=\textwidth]{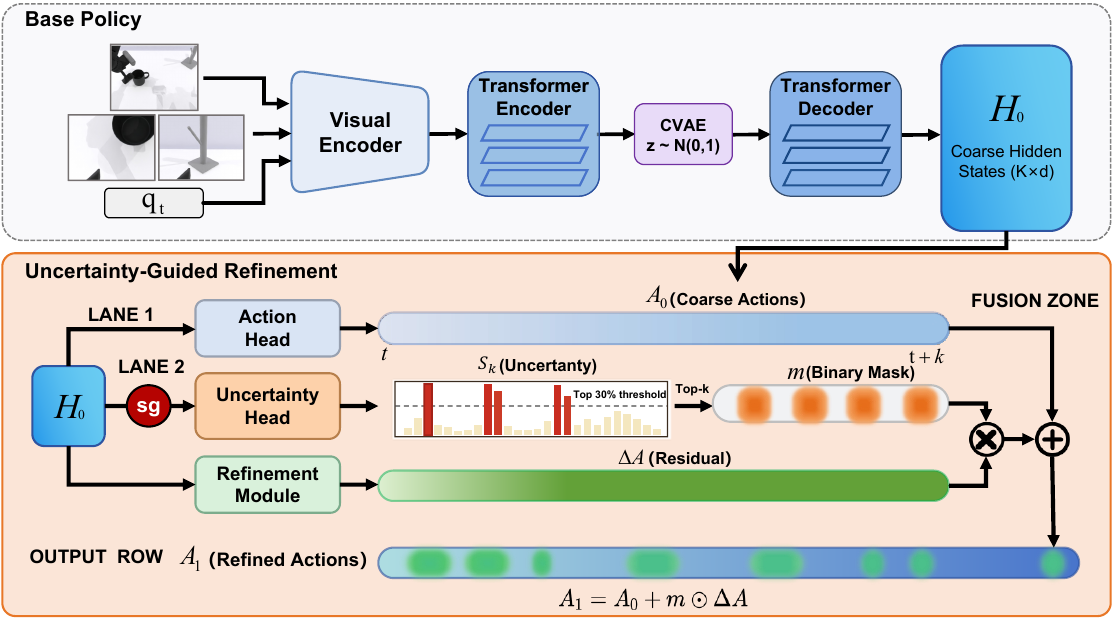}
\caption{Overview of the proposed uncertainty-guided sparse refinement pipeline. ACT first predicts a coarse action chunk and the corresponding decoder hidden states. A lightweight uncertainty head reads the detached coarse hidden states and predicts one scalar uncertainty score for each timestep. The top-$\lfloor \rho K \rfloor$ timesteps are selected to form a binary refinement mask. Residual correction is applied only to the selected timesteps, while the remaining timesteps keep the coarse prediction unchanged.}
\label{fig:overview}
\end{figure*}

\section{RELATED WORK}
\textbf{Chunk-based visuomotor policies.}
Visuomotor imitation learning trains policies to map observations to actions from expert demonstrations~\cite{mandlekar2022matters,florence2022ibc}. Chunk-based action prediction has become a dominant paradigm by predicting multiple future actions jointly, improving temporal consistency and reducing compounding errors~\cite{ross2011reduction}. ACT~\cite{zhao2023act} demonstrates that CVAE-based chunk decoding is effective for bimanual manipulation and has been extended to mobile settings in Mobile ALOHA~\cite{fu2024mobile}. Diffusion Policy~\cite{chi2025diffusion} formulates action prediction as iterative denoising over action chunks, while Behavior Transformer~\cite{shafiullah2022bet}, VQ-BeT~\cite{lee2024vqbet}, and HYDRA~\cite{belkhale2023hydra} explore alternative temporal or latent action abstractions. Broader high-capacity robot policies such as PerAct~\cite{shridhar2023peract}, RT-1~\cite{brohan2023rt1}, Octo~\cite{ghosh2024octo}, and OpenVLA~\cite{kim2024openvla} further push policy learning toward multi-task and large-data settings. RoboMimic~\cite{mandlekar2022matters}, MimicGen~\cite{mandlekar2023mimicgen}, CALVIN~\cite{mees2022calvin}, LIBERO~\cite{liu2023libero}, and RoboTwin~\cite{mu2024robotwin} provide datasets and benchmarks for reproducible manipulation evaluation. Despite these advances, existing methods focus on improving overall chunk quality and do not explicitly target failures that concentrate on a few critical timesteps.
\textbf{Refinement and uncertainty in action prediction.}
Coarse-to-fine and iterative refinement strategies have been studied in sequence prediction and robot control. Residual policy learning~\cite{silver2018residual} and residual reinforcement learning~\cite{johannink2019residual} add corrective actions on top of a base policy, while Diffusion Policy performs multi-step refinement through iterative denoising. Robust imitation learning methods such as DART~\cite{laskey2017dart} reduce covariate shift by injecting noise into demonstrations so that policies learn to recover from errors. However, these methods refine or regularize behavior globally without distinguishing easy and difficult timesteps within a chunk. Uncertainty estimation has also been explored via heteroscedastic regression~\cite{kendall2017uncertainty}, deep ensembles~\cite{lakshminarayanan2017simple}, and MC-Dropout~\cite{gal2016dropout}. In robot learning, uncertainty has also supported learning from corrections~\cite{losey2018uncertainty} and runtime failure detection for imitation policies~\cite{xu2025failure}. Our method instead uses temporal uncertainty to determine \emph{where} within an action chunk sparse residual correction should be applied.

\section{METHOD}

\subsection{Preliminaries}
We build our method on top of the Action Chunking Transformer (ACT) policy. At each timestep $t$, the policy takes as input three RGB observations and the robot state $q_t$, and predicts a future action chunk of length $K$. In our setting, the three camera views are \texttt{cam\_high}, \texttt{cam\_right\_wrist}, and \texttt{cam\_left\_wrist}. We use a chunk size of $K=50$, and both the action and state dimensions are 14. Given the observation at timestep $t$, the ACT policy predicts a future action sequence
\begin{equation}
A = [a_t, a_{t+1}, \dots, a_{t+K-1}].
\end{equation}

\subsection{Uncertainty-Guided Sparse Refinement}
\subsubsection{Motivation}
As discussed in Section~\ref{sec:intro}, ACT failures can arise from \emph{temporally localized errors} at critical transitions (contact, alignment, handover) rather than uniformly poor predictions. This motivates a refinement strategy that focuses on difficult local regions instead of correcting the entire chunk uniformly.
\subsubsection{Coarse-to-Refine Chunk Prediction}
To address this issue, we adopt a coarse-to-refine design. Given the input observation at timestep $t$, the ACT decoder first produces a coarse action chunk prediction
\begin{equation}
A_0 = [a_t^{(0)}, a_{t+1}^{(0)}, \dots, a_{t+K-1}^{(0)}].
\end{equation}
A refinement module then predicts a full-length residual $\Delta A \in \mathbb{R}^{K \times d_a}$, where $d_a$ is the action dimension. A binary refinement mask $m \in \{0,1\}^{K}$ selects which timesteps receive correction:
\begin{equation}
A_1 = A_0 + m \odot \Delta A,
\end{equation}
where $\odot$ denotes element-wise multiplication broadcast across the action dimension. The choice of $m$ distinguishes the compared variants:
\begin{itemize}
    \item \textbf{Full refinement} ($m = \mathbf{1}$): all timesteps are refined.
    \item \textbf{Block refinement}: $m_k = 1$ for a contiguous block of length $B$ at a randomly sampled starting position, and $m_k = 0$ elsewhere.
    \item \textbf{Uncertainty-guided refinement} (proposed): $m_k = 1$ for the top-$\lfloor \rho K \rfloor$ timesteps ranked by predicted uncertainty, and $m_k = 0$ otherwise.
\end{itemize}
This unified formulation makes clear that all refinement variants share the same coarse-to-refine architecture and differ only in how the mask $m$ is constructed. The full refinement variant spreads correction uniformly; block refinement concentrates it on a random local segment; and our method allocates it in a data-dependent manner guided by temporal uncertainty.
\subsubsection{Sparse Block Refinement}
Full-chunk refinement treats all timesteps equally, even though many are already predicted well. This may spread the correction budget too thinly and introduce unnecessary perturbations to already-correct actions. As a sparse alternative, block refinement sets $m_k = 1$ for a contiguous block of length $B{=}12$ within each 50-step action chunk (i.e., 24\% of timesteps). The block starting position is sampled uniformly at random during training. This design provides a stronger inductive bias for correcting temporally localized errors than full-chunk refinement, but remains position-agnostic at test time because the block location does not depend on the input.
\subsubsection{Uncertainty-Guided Sparse Refinement}
While block refinement improves locality, a randomly chosen block remains position-agnostic: it does not know which timesteps are truly difficult for the current input. To address this limitation, we propose uncertainty-guided sparse refinement.

We attach a lightweight uncertainty head (a single linear layer) to the coarse decoder hidden states $H_0 \in \mathbb{R}^{K \times d}$ and predict a scalar log-variance $s_k \in \mathbb{R}$ for each timestep $k$:
\begin{equation}
s_k = \text{Linear}(\text{sg}(h_k^{(0)})), \quad k = 0, \dots, K{-}1,
\end{equation}
where $\text{sg}(\cdot)$ denotes stop-gradient (see Section~\ref{sec:decoupled_uncertainty}) and $h_k^{(0)}$ is the $k$-th coarse hidden state. The refinement mask is then constructed by selecting the top-$\lfloor \rho K \rfloor$ timesteps ranked by $s_k$:
\begin{equation}
m_k = \mathbf{1}\!\left[s_k \geq s_{(\lceil (1-\rho)K \rceil)}\right],
\end{equation}
where $s_{(j)}$ denotes the $j$-th order statistic (sorted in ascending order) and $\rho = 0.3$ is the refinement ratio. In this way, the refinement budget is allocated in a data-dependent manner according to the temporal difficulty of the sample.

We set the refinement ratio to $\rho=0.3$ and the block length to $B=12$ based on preliminary validation, and keep them fixed across all tasks to avoid task-specific tuning. With the default chunk length $K=50$, $B=12$ refines 24\% of a chunk, placing the block-refinement baseline in a similar sparse-refinement budget regime to the uncertainty-guided setting.

Operationally, the uncertainty head only ranks timestep-level refinement difficulty; the selected top-$\lfloor \rho K \rfloor$ positions are refined by the residual branch, whereas unselected positions are copied directly from the coarse chunk.

We use a scalar uncertainty per timestep rather than joint-wise uncertainty because the purpose of the uncertainty branch is temporal step selection, not per-joint reweighting. This design is lightweight, stable under limited demonstration data, and naturally matches the top-$k$ selection mechanism.

The additional overhead is modest because the uncertainty head is implemented as a small linear projection on top of the existing decoder states, and the top-$k$ selection is performed over $K$ timestep-level uncertainty scores without an additional visual encoding pass. The refinement branch reuses the shared ACT representation, and residual correction is applied only at the selected timesteps.

\begin{figure*}[t]
\centering
\includegraphics[width=0.98\textwidth]{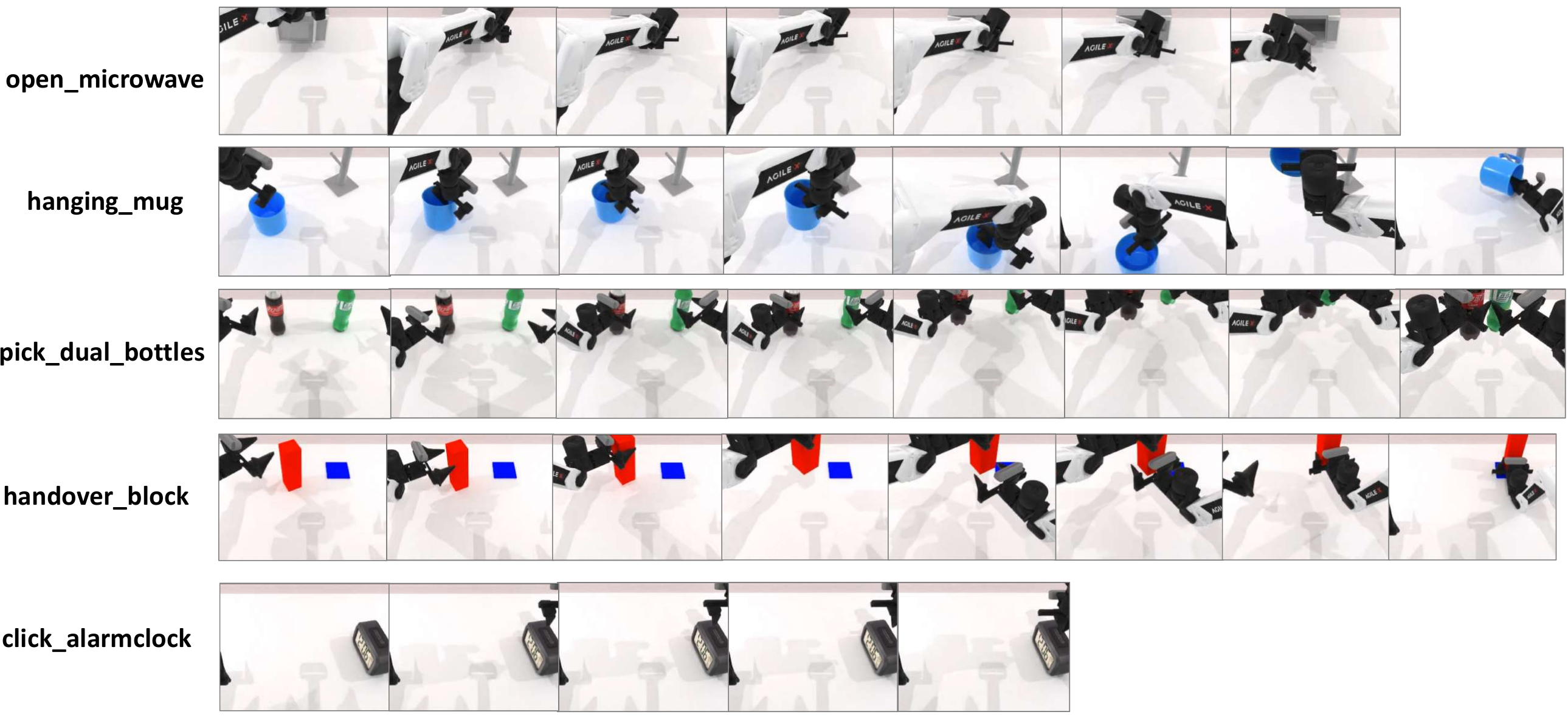}
\caption{Representative simulation rollouts from the five RoboTwin manipulation tasks used in our evaluation. Each row shows frames from one task: \textit{open\_microwave}, \textit{hanging\_mug}, \textit{pick\_dual\_bottles}, \textit{handover\_block}, and \textit{click\_alarmclock}.}
\label{fig:simulation}
\end{figure*}

\subsubsection{Decoupled Uncertainty Learning}
\label{sec:decoupled_uncertainty}
To keep the role of uncertainty estimation interpretable, we prevent the uncertainty objective from directly updating the coarse action branch. Specifically, the uncertainty head operates on detached coarse hidden states, and the uncertainty loss is computed from detached coarse predictions. As a result, the uncertainty branch serves as an auxiliary estimator for selecting refinement locations, rather than acting as an additional hidden supervision path for the coarse predictor. This decoupled design makes the benefit of our method easier to attribute to uncertainty-guided refinement itself.

\subsubsection{Training Objective}
The overall training loss consists of four terms. The \textbf{coarse prediction loss} supervises the initial action chunk using mean squared error:
\begin{equation}
\mathcal{L}_{\text{coarse}} = \frac{1}{K}\sum_{k=0}^{K-1} \| a_{t+k}^{(0)} - a_{t+k}^{*} \|^2,
\end{equation}
where $a_{t+k}^{*}$ denotes the ground-truth action at timestep $t{+}k$. The \textbf{refinement loss} supervises the corrected actions at the selected timesteps:
\begin{equation}
\mathcal{L}_{\text{refine}} = \frac{1}{|\mathcal{S}|}\sum_{k \in \mathcal{S}} \| a_{t+k}^{(1)} - a_{t+k}^{*} \|^2,
\end{equation}
where $\mathcal{S} = \{k : m_k = 1\}$ is the set of timesteps selected by the refinement mask, and $a_{t+k}^{(1)} = a_{t+k}^{(0)} + \Delta a_{t+k}$ is the refined prediction. The \textbf{uncertainty loss} follows the heteroscedastic regression formulation~\cite{kendall2017uncertainty}. The key idea is to model each coarse prediction error as drawn from a Gaussian with learned variance $\exp(s_k)$, yielding a negative log-likelihood objective:
\begin{equation}
\mathcal{L}_{\text{uncert}} = \frac{1}{K}\sum_{k=0}^{K-1} \left( \frac{\| \bar{a}_{t+k}^{(0)} - a_{t+k}^{*} \|^2}{\exp(s_k)} + s_k \right),
\end{equation}
where $s_k$ is the predicted log-variance for timestep $t{+}k$, and $\bar{a}_{t+k}^{(0)} = \text{sg}(a_{t+k}^{(0)})$ denotes the coarse prediction with gradients detached. This loss encourages $s_k$ to be large where the coarse error is large, providing a learned difficulty signal without back-propagating through the coarse branch. Following ACT, we also include a \textbf{KL divergence} term $\mathcal{L}_{\text{KL}} = D_{\text{KL}}(q(z|A^*) \| p(z))$ that regularizes the CVAE posterior toward the unit Gaussian prior. The total loss is:
\begin{equation}
\mathcal{L} = \mathcal{L}_{\text{coarse}} + \mathcal{L}_{\text{refine}} + \lambda_u \mathcal{L}_{\text{uncert}} + \lambda_{\text{KL}} \mathcal{L}_{\text{KL}},
\end{equation}
where $\lambda_u$ and $\lambda_{\text{KL}}$ are weighting coefficients (set to $\lambda_u{=}1$ and $\lambda_{\text{KL}}{=}10$ in all experiments).

\section{Simulation Experiments}

\subsection{Benchmarks and Implementation Details}
We evaluate all methods on five RoboTwin manipulation tasks: \textit{handover\_block}, \textit{pick\_dual\_bottles}, \textit{hanging\_mug}, \textit{open\_microwave}, and \textit{click\_alarmclock}. Figure~\ref{fig:simulation} visualizes representative simulation rollouts from these tasks. For all tasks, we use the \textit{demo\_clean} setting with 50 demonstrations per task. The processed ACT dataset stores the action trajectory, robot state, and three RGB camera streams, namely \texttt{cam\_high}, \texttt{cam\_right\_wrist}, and \texttt{cam\_left\_wrist}. Following the ACT data pipeline, each task is split at the episode level into 80\% training and 20\% validation sets.

\begin{table*}[t]
\centering
\caption{Success rate (\%) of different methods on five RoboTwin manipulation tasks. Best results per task are highlighted in bold.}
\label{tab:main_result}
\small
\begin{tabular}{lcccccc}
\toprule
Method  & Handover & Pick Bottles & Hang Mug & Open Micro. & Click Alarm. & Overall \\
\midrule
ACT \cite{zhao2023act} & 45\% & 20\% & 15\%  & 65\% & 29\% & 34.8\% \\
DP \cite{chi2025diffusion} & 10\% & 24\% & 8\%  & 5\% & \textbf{61\%} & 21.6\% \\
UGR & \textbf{50\%} & \textbf{27\%} & \textbf{21\%}  & \textbf{78\%} & 41\% & \textbf{43.4\%} \\
\bottomrule
\end{tabular}
\end{table*}

\begin{table}[t]
\centering
\caption{Ablation study on the Click Alarm task.}
\label{tab:ablation}
\footnotesize
\resizebox{\columnwidth}{!}{%
\begin{tabular}{llc}
\toprule
Variant & Refinement Strategy & Success Rate \\
\midrule
act\_base & None  & 29\% \\
act\_cmask & None + causal decoder mask & 24\% \\
act\_refine\_full & Full-chunk residual  & 31\% \\
act\_refine\_block & Sparse block & 28\% \\
act\_refine\_uncert & Uncertainty-guided & \textbf{41\%} \\
\bottomrule
\end{tabular}
}
\end{table}

All ACT-family variants use the same ACT backbone with a ResNet-18 visual encoder and a Transformer of hidden dimension 512, feedforward dimension 3200, 4 encoder layers, 7 decoder layers, and 8 attention heads. The action chunk length is fixed to 50, and both the action and state dimensions are 14. We train these models using AdamW with batch size 8, learning rate $10^{-5}$, weight decay $10^{-4}$, and KL weight 10 for 6000 epochs. All methods use the same task split, evaluation seed set, and checkpoint selection protocol based on \textit{policy\_best.ckpt}. ACT-based policies use temporal aggregation during evaluation.
\subsubsection{\textbf{Baselines}}
We compare our method with both representative policy baselines and ACT-family variants. For the main performance comparison in Table~\ref{tab:main_result}, we report \textbf{ACT}, the original ACT baseline; \textbf{DP}, Diffusion Policy; and \textbf{UGR}, our uncertainty-guided refinement model. For ablation within the ACT family, we further consider three intermediate variants. \textbf{act\_cmask} adds a causal self-attention mask to the ACT decoder while keeping the rest of the architecture unchanged. \textbf{act\_refine\_full} introduces coarse-to-refine prediction with full-chunk residual refinement. \textbf{act\_refine\_block} further restricts refinement to a contiguous temporal block of length 12 within each 50-step action chunk, providing a sparse local refinement baseline. The final ACT-family model, \textbf{act\_refine\_uncert}, replaces position-agnostic sparse refinement with uncertainty-guided sparse refinement by selecting the top 30\% most uncertain timesteps for correction.
\subsubsection{\textbf{Evaluation Metrics}}
Our primary evaluation metric is task success rate. For each task, a rollout is counted as successful if the environment-defined success condition is satisfied before the maximum task horizon is reached; otherwise it is counted as a failure. All methods are evaluated under the same seed protocol to ensure fair comparison.
During evaluation, ACT-based policies use temporal aggregation rather than directly executing a single raw chunk prediction from one query. At each environment step, the policy is queried repeatedly and the final executed action is obtained by aggregating the available action predictions over time. All methods are evaluated over the same set of evaluation episodes per task. For the ACT-family variants, training uses the same random seed and results are reported from the best checkpoint selected by validation loss. 
\subsection{Performance Comparison}
Table~\ref{tab:main_result} reports the main comparison among ACT, Diffusion Policy (DP), and our uncertainty-guided refinement model (UGR) on five RoboTwin manipulation tasks.

In the single-seed results, UGR achieves the best success rate on four of the five tasks. Compared with ACT, it improves performance on all five tasks, with gains ranging from $+5\%$ to $+13\%$ (Table~\ref{tab:main_result}). Compared with DP, UGR performs better on four tasks and trails only on \textit{click\_alarmclock}. These results are consistent with our hypothesis that data-dependent refinement targeting uncertain timesteps can improve action chunk prediction.

\subsection{Ablation Studies}
Table~\ref{tab:ablation} compares five ACT variants on the Click Alarm task, each using a different refinement strategy. The comparison is designed to isolate the effect of refinement strategy while keeping the backbone architecture fixed.

\texttt{act\_cmask} does not improve over \texttt{act\_base} (24\% vs. 29\%), indicating that decoder causality alone is insufficient. Full-chunk refinement (\texttt{act\_refine\_full}, 31\%) slightly improves over the baseline, confirming the benefit of coarse-to-refine prediction. Sparse block refinement (\texttt{act\_refine\_block}, 28\%) still does not outperform full refinement, suggesting that a randomly placed block does not reliably cover the most error-prone timesteps. Uncertainty-guided refinement (\texttt{act\_refine\_uncert}, 41\%) achieves the best result by a clear margin, providing evidence that data-dependent timestep selection is more effective than position-agnostic alternatives.

\section{Conclusion}
In this paper, we argue that ACT failures in long-horizon manipulation often arise from temporally localized errors rather than uniformly poor action chunks. We therefore propose an uncertainty-guided sparse refinement framework that uses step-wise uncertainty to apply residual correction only at the most uncertain timesteps while keeping the uncertainty branch decoupled from the coarse predictor. On five RoboTwin tasks, \texttt{act\_refine\_uncert} improves or remains competitive with representative baselines and position-agnostic refinement variants, suggesting that targeted correction of critical local transitions is a promising direction for chunk-based manipulation policies.

\textbf{Limitations.} Our study is limited to simulation and uses a simple scalar uncertainty head with fixed sparse-refinement hyperparameters.
\bibliographystyle{IEEEbib}
\IEEEtriggeratref{18}
\bibliography{ref}

\end{document}